\documentclass{article}
\usepackage[margin=2.5cm]{geometry}
 
\usepackage{cite}
\usepackage{amsfonts,amsmath,amsthm}
\usepackage{algorithmic}
\usepackage{algorithm}

\usepackage{graphicx}
\DeclareGraphicsExtensions{.pdf,.png,.jpg,.jpeg }

\theoremstyle{definition}
\newtheorem{definition} {Definition} 
\newtheorem{example} {Example} 
\newtheorem{remark} {Remark} 
 
\begin{document}
\title{Geiringer Theorems: From Population Genetics to Computational Intelligence, Memory Evolutive Systems and Hebbian Learning
}

\author{Boris~S.~Mitavskiy \thanks{Email: bom4@aber.ac.uk }\\  Department of Computer Science\\
  Aberystwyth University, 
  Aberystwyth SY23 3DB, UK   \and Elio~Tuci  \\ Department of Computer Science\\
  Aberystwyth University,   Aberystwyth SY23 3DB, UK \and Chris~Cannings \\ School of Mathematics and Statistics\\
  University of Sheffield, 
  Sheffield, England, S10 2RX \and Jonathan~Rowe \\ School of Computer Science\\
  University of Birmingham, 
  Edgbaston, Birmingham, B15 2TT, United Kingdom \and Jun~He \\   Department of Computer Science\\
  Aberystwyth University, 
  Aberystwyth SY23 3DB, UK}
  \date{}
\maketitle

\begin{abstract}
The classical Geiringer theorem addresses the limiting frequency of occurrence of various alleles after repeated application of crossover. It has been adopted to the setting of evolutionary algorithms and, a lot more recently, reinforcement learning and Monte-Carlo tree search methodology to cope with a rather challenging question of action evaluation at the chance nodes. The theorem motivates  novel dynamic parallel algorithms that are explicitly described in the current paper for the first time. The algorithms involve independent agents traversing a dynamically constructed directed graph that possibly has loops. A rather elegant and profound category-theoretic model of cognition in biological neural networks developed by a well-known French mathematician, professor Andree Ehresmann jointly with a neurosurgeon, Jan Paul Vanbremeersch over the last thirty years provides a hint at the connection between such algorithms and Hebbian learning.

\end{abstract}

\section{Introduction}\label{introSect}
The hereditary material of individuals is arranged along structures
called chromosomes which may be linear or circular. The latter of
these are found in bacteria and we focus here only on the linear
case. Within a species each individual has a number of copies of
each of a number of homologous (i.e. matching) chromosomes. Thus,
for example, humans have 22 homologous pairs (called diploid) while
many organisms exist with four (tetraploid), six (hexaploid) or
eight (octaploid)chromosomes. Less commonly there may be an odd
number of chromosomes.

The implication of the hereditary units (genes) being arranged along
chromosomes is that Mendel's 2nd Law (that distinct characters are
transmitted independently) is violated. For ease we consider a
diploid individual and suppose we have a particular homologous pair
of chromosomes and some $n$ genes arranged along that chromosomes.
During the formation of germ cells (meiosis) a new chromosome is
formed from the pair. If we denote the genes along the chromosome of
the individual obtained from its mother as $\{0,0,\ldots,0\}$ and
from its father as $\{1,1,\ldots,1\}$ (where each vector is of
length $n$) then the chromosome formed will be some binary string
e.g. $l=\{0,1,1,\ldots,0,1\}$ with some associated probability. Thus
the genes are recombined i.e. mixed from the grandparents while the
order is (generally) conserved. Further the genes will take values
referred to as alleles, and the alleles along the chromosomes will
be transmitted to the germ cell unchanged
unless some  mutation occurs.

There are two important features of the recombination process. The
first is that this structure creates correlations between the
transmission of genes which are, in some sense, close to each other
along the chromosome. This has both the potential to allow blocks of
genes to be inherited together as recombination is suppressed in
certain regions, and also to rearrange alleles into potentially
evolutionarily advantageous combinations. We can also potentially
infer the close linkage (position) of genes through the detection of
correlations. There is however, as proved in \cite{GeirOrigion}, a
steady weakening of these correlations through time and the
asymptotic mutual independence of the allele frequencies, in the
absence of selection. Thus the correlations can only be used for a
limited time from founder population in which a correlation
fortuitously
existed. Quite a while later \cite{SinclRabinov} established bounds on the actual rates of convergence towards the independence of the allele frequencies.  

The idea of using Geiringer theorem in evolutionary computation with the aim of predicting the outcome of repeated applications of crossover has been introduced in \cite{GeiringerMuhlenbeim} and motivated the development of EDAs (estimation of distribution algorithms). The classical version of the theorem is stated in terms of a discrete-time quadratic system ``infinite population" model very much as follows: let $\Omega =
\prod_{i=1}^n A_i$ denote the search space of a given genetic
algorithm. (Intuitively, $n$ is the number of loci and $A_i$ is
the set of alleles corresponding to the $i$th gene.)
Denote by $\Lambda$ the collection of all probability
distributions on $\Omega$. Now fix a probability distribution
$\lambda \in \Lambda$ and consider the sequence of probability
distributions $\lambda, \, C(\lambda), \, C^2(\lambda), \ldots,
C^n(\lambda), \ldots$ where $\mathcal{C}(p)(k) = \sum_{i, j} p(i)
p(j) r_{(i, \, j \rightarrow k)}$ and $r_{(i, \, j \rightarrow
k)}$ denotes the probability of obtaining the individual $k$ from
the parents $i$ and $j$ after crossover. Here crossover can be
thought of as an operator which takes a pair of elements of the
search space (the parents) and produces another element of the
search space (the child) by mingling the alleles of the parents.
This will be discussed in more detail later in the paper. Denote
by $\lambda_i$ the marginal distribution of $\lambda$ on $A_i$.
The classical Geiringer theorem says that $\lim_{n \rightarrow
\infty} C^n(\lambda) \rightarrow \prod_{i=1}^n \lambda_i$ (meaning
that the frequency of occurrence of an individual $\mathbf{x} =
(x_1, \, x_2, \ldots, x_n) \in \Omega$ under the limiting
distribution $\lim_{n \rightarrow \infty} C^n(\lambda)$ is just
the product of the frequencies of $x_i$ under the distributions
$\lambda_i$). In \cite{GPGeir} this theorem has been generalized
to cover the cases of variable-length GA's and homologous linear
GP crossover. The limiting distributions of the frequency of
occurrence of individuals belonging to a certain schema under
these algorithms have been computed. An alternative approach that provides estimates on the convergence rates towards the limiting distribution
of the quadratic dynamical system $\{C^n\}_{n=1}^{\infty}$ is provided in \cite{SinclRabinov}. A rather general and powerful finite-population version of Geiringer theorem
based on the unique uniform stationary distribution of the Markov chain of populations on the orbit of the ``joint" group action of bijective recombination transformations has been established in \cite{MitavGeir}. Knowing that the stationary distribution of this Markov chain is uniform, for the most types of traditional GAs and GP with homologous recombination it is quite routine to derive the the limiting frequency of occurrence of various schemata and, not in vain, the corresponding formulas come out exactly the same as in the infinite-population model. Intuitively speaking, this happens due to the fact that as the sample size gets larger and larger, sampling with replacement (i.e. binomial sampling) approaches in distribution sampling without replacement (i.e. hypergeometric sampling). A rigorous argument does require careful analysis and will be presented in a sequel research paper. Apart from the already established infinite-population versions of the Geiringer theorems, the methodology for deriving such theorems allows one to obtain a formula for the limiting frequency of occurrence of various schemata for non-linear GP with homologous recombination (see \cite{MitavRowGeirGenProgr}) without much effort. Gieringer-like theorems for non-homologous recombination remained an open question until recently, see \cite{MitavRowGeirNewMain}, when such a theorem has been derived in the setting of POMDPs (partially observable Markov decision processes) and model-free reinforcement learning with the aim of enhancing action evaluation at the chance nodes. The technical challenges have been overcome thanks to a lovely application of the classical and extremely useful triviality known as the Markov inequality (see, for instance, \cite{DoerrB}) and enhancing the methodology for deriving the Geiringer-like results in \cite{MitavRowGeirMain} through exploiting a slightly extended lumping quotients of Markov chains technique that has been successfully developed, improved and exploited to estimate stationary distributions of Markov chains modeling EAs in a series of articles: \cite{MitavJohnChrisLotharConf}, \cite{MitavJonAldenLotharGPEM}, \cite{MitavChrisConf} and \cite{MitavChris}. Finally, the later version of the theorem has been further generalized to allow recombination over arbitrary set covers, rather than being limited to equivalence relations using the particular case in \cite{MitavRowGeirNewMain} in conjunction with the tools mentioned above in \cite{MitavJunGeirFoga2013}. While the original purpose of the last two finite-population Geiriger-like theorems is taking advantage of the intrinsic similarities within the state-action set encountered by a learning agent to evaluate actions with the aim of selecting an optimal one, the parallel algorithms on the evolving digraph where the nodes are states and actions are edges from a current state to the one obtained upon executing an action, that are motivated by the theorems exhibit similarities to the way Hebbian learning takes place in biological neural networks that is further supported by Andree Ehresmann's category-theoretic model of cognitive processes called a ``Memory Evolutive System" (see, for instance, \cite{ehresmannBaasVanbremeersch2004}, \cite{ehresmannSmeonov2012}, \cite{EhresmannVanbremeersch2006}, \cite{EhresmannVanbremeersch2007Book} and many more related articles). In the current article we explain the new theorems as well as the algorithms they motivate, how such algorithms may fit the Memory Evolutive Systems model that hopefully provides a deeper understanding of cognitive processes and makes a further step in the development of intelligent computational systems.
\section{Mathematical Framework, Statement of the new Geiringer-like Theorem for Action Evaluation in POMDPs and the Corresponding Action-Evaluation Algorithms.}\label{FrameworkAndAlgorithmSect}
\begin{figure}[htp]
\centerline{\includegraphics[height=10cm]{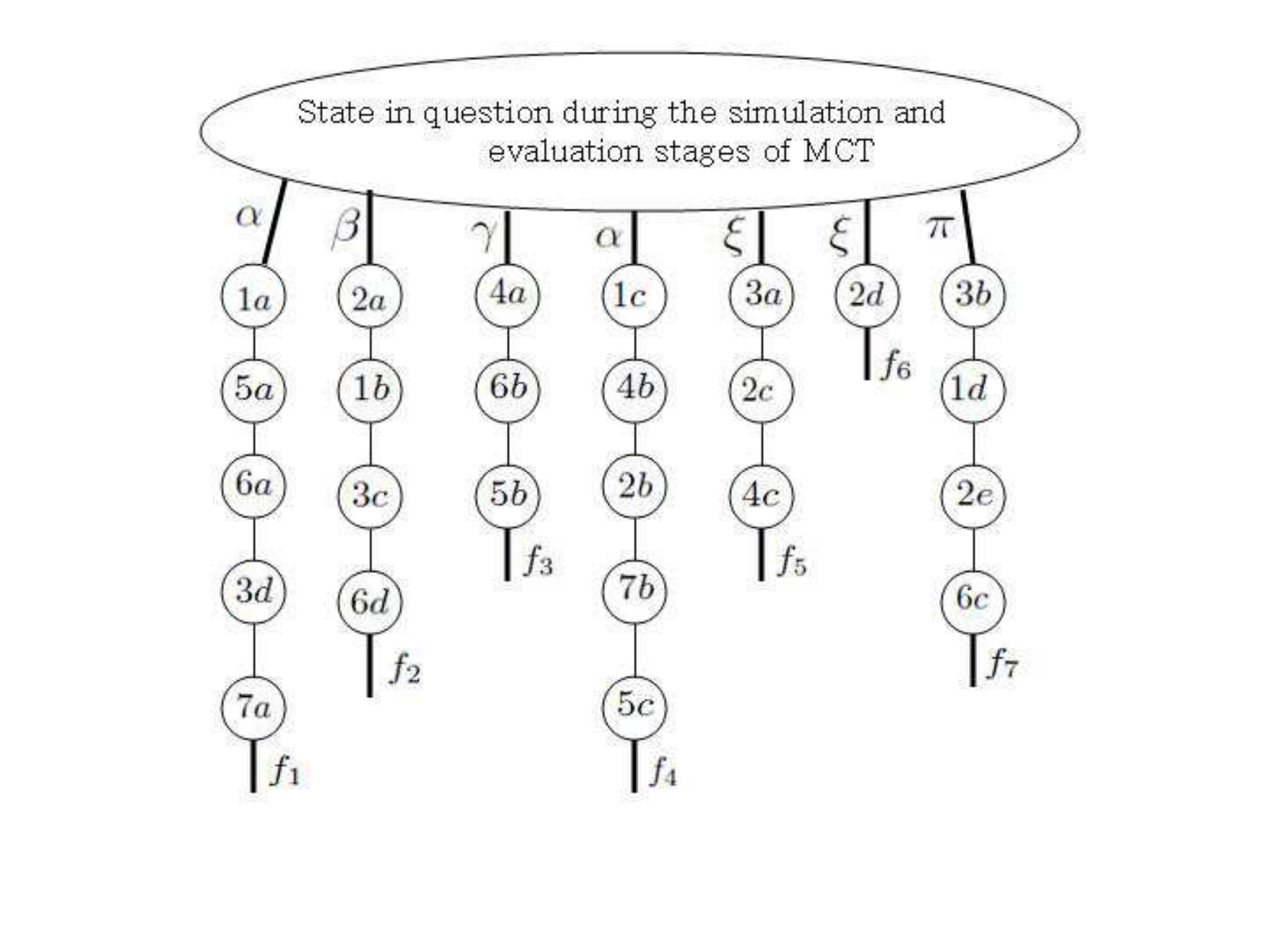}}
\caption{An example of a population consisting of seven rollouts. Equivalence classes of states are denoted by distinct numbers so that the letters written next to these numbers distinguish the individual states as in remark~\ref{crossoverConvRepresRem}. Distinct actions under evaluation (see definition~\ref{treeRootedByChanceNode}) are denoted by different letters of Greek alphabet.}
\label{PopOfRollsFig}
\end{figure}

A great number of questions in machine learning, computer game intelligence, control theory, and in cognitive processes in general involve decision-making by an agent under a specified set of circumstances. In the most general setting, the problem can be described mathematically in terms of the state and action pairs as follows. A state-action pair is an ordered pair of the form $(s, \, \vec{\alpha})$ where $\vec{\alpha} = \{\alpha_1, \, \alpha_2, \ldots, \alpha_n\}$ is the set of actions (or moves, in case the agent is playing a game, for instance) that the agent is capable of taking when it is in the state (or, in case of a game, a state might be sometimes referred to as a position) $s$. Due to randomness, hidden features, lack of memory, limitation of the sensor capabilities etc, the state may be only partially observable by the agent. Mathematically this means that there is a function $\phi: S \rightarrow O$ (as a matter of fact, a random variable with respect the unknown probability space structure on the set $S$) where $S$ is the set of all states which could be either finite or infinite while $O$ is the set (usually finite due to memory limitations) of observations having the property that whenever $\phi(s_1) = \phi(s_2)$ (i.e. whenever the agent can not distinguish states $s_1$ and $s_2$) then the corresponding state action pairs $(s_1, \, \vec{\alpha})$ and $(s_2, \, \vec{\beta})$ are such that $\vec{\alpha} = \vec{\beta}$ (i.e. the agent knows which actions it can possibly take based only on the observation it makes). The general problem of reinforcement learning is to decide which action is best suited given the agent's knowledge (that is the observation that the agent has made as well as the agent's past experience). In case when immediate payoffs are not known, self-simulated random trials, named \emph{rollouts} by the Monte-Carlo tree search community, until the terminal state (or until a payoff value is available) are implemented and, based on a sufficiently large sample of such rollouts, an agent has to decide which actions are most advantageous in which states. Under a reasonable assumption that adaptation and unsupervised learning in biological organisms takes place largely based on trial followed by reward mechanisms, the corresponding learning models are very much similar. Thus, suppose a sequence of independent rollouts has been simulated or, in case of adoptive learning agents, such as robots or biological systems, gained via some learning experience. Due to alterations within competitor's, opponent's or enemy's strategies and unavailability of immediate feedback, reward or payoff, the combinatorial explosion in the number of possibilities that could take place after an execution of the same action at a given state, even in case when the learning agent has just several actions to choose from at a given state, is tremendous. The situation is significantly exacerbated due to the unknown information and randomness within the environment, that is likely to cause drastic differences in the eventual payoffs after the same action starting at a specified observable state has been taken. At the same time, only a limited sample of rollouts (or trials) is available for action evaluation. Needless to say, optimal action evaluation under such circumstances is an extremely complex and challenging task. While the unfortunate reality that the total number of observable states is often drastically limited comparing to the total overall number of states (i.e. when the non-observable information is taken into account), contributes to the overall complexity, it turns out, that this could also be exploited to design more effective action-evaluation policies and the design of such policies is motivated by the Geiringer-like theorems. For the sake of completeness, we now explain these ideas and state the Geiringer-like theorem established in \cite{MitavRowGeirNewMain} in a mathematically rigorous fashion. Let $S$ denote the set of states (enormous but finite in this framework). Formally each state $\vec{s} \in S$ is an ordered pair $(s, \vec{\alpha})$ where $\vec{\alpha}$ is the set of actions an agent can possibly take when in the state $\vec{s}$. Let $\sim$ be an equivalence relation on $S$. Without loss of generality we will denote every equivalence class by an integer $1, \, 2, \ldots, i, \ldots, \in \mathbb{N}$ so that each element of $S$ as an ordered pair $(i, \, a)$ where $i \in \mathbb{N}$ and $a \in A$ with $A$ being some finite alphabet. With this notation $(i, \, a) \sim (j, \, b)$ iff $i=j$. Intuitively, $S$ is the set of all possible states and $\sim$ is the similarity relation on $S$ where two states are equivalent under $\sim$ if and only if they contain identical observable information.\footnote{Other scenarios where states might be equivalent due to obvious observable similarity are also conceivable and certainly play an important role in Monte-Carlo tree search methodology.} We will also require that for two equivalent states $\vec{s_1} = \{s_1, \vec{\alpha}_1\}$ and $\vec{s_2} = \{s_2, \vec{\alpha}_2\}$ under $\sim$ there are bijections $f_1: \vec{\alpha}_1 \rightarrow \vec{\alpha}_2$ and $f_2: \vec{\alpha}_2 \rightarrow \vec{\alpha}_1$. For the time being, these bijections should be obvious and natural from the representation of the environment (and actions) and reflect the similarity between these actions.
\begin{definition}\label{treeRootedByChanceNode}
Suppose we are given a state $\vec{s} = (s, \vec{\alpha})$ and a sequence $\{\alpha_i\}_{i=1}^b$ of actions in $\vec{\alpha}$ (it is possible that $\alpha_i = \alpha_j$ for $i \neq j$). We may then call $\vec{s}$ a \emph{root state}, or a \emph{state in question}, the sequence $\{\alpha_i\}_{i=1}^b$, the \emph{sequence of actions under evaluation} and the set of actions $\mathcal{A} = \{\alpha \, | \, \alpha = \alpha_i$ for some $i$ with $1 \leq i \leq b\}$, the set of actions under evaluation.
\end{definition}
\begin{definition}\label{RolloutDefn}
A \emph{rollout} with respect to the state in question $\vec{s} = (s, \vec{\alpha})$ and an action $\alpha \in \vec{\alpha}$ is a sequence of states following the action $\alpha$ and ending with a terminal label $f \in \Sigma$ where $\Sigma$ is an arbitrary set of labels\footnote{Intuitively, each terminal label in the set $\Sigma$ represents a terminal state that we can assign a numerical value to via a function $\phi: \, \Sigma \rightarrow \mathbb{Q}$. The reason we introduce the set $\Sigma$ of formal labels as opposed to requiring that each terminal label is a rational number straight away, is to avoid confusion in the upcoming definitions}, which looks as $\{(\alpha, \, s_1, \, s_2, \ldots, s_{t-1}, \, f)\}$. For technical reasons that will become obvious later we will also require that $s_i \neq s_j$ for $i \neq j$ (it is possible and common to have $s_i \sim s_j$ though). We will say that the total number of states in a rollout (which is $k-1$ in the notation of this definition) is the \emph{height} of the rollout.
\end{definition}
\begin{remark}\label{crossoverConvRepresRem}
In the paragraph preceding definition~\ref{treeRootedByChanceNode} we have introduced a convenient notation for states to emphasize their respective equivalence classes. With such notation a typical rollout would appear as a sequence $\{(\alpha, \, (i_1, \, a_1), \, (i_2, \, a_2), \ldots, (i_{t-1}, a_{t-1}), \, f)\}$ with $i_j \in \mathbb{N}$ while $a_i \in A$. According to the requirement in definition~\ref{RolloutDefn}, $i_j = i_k$ for $j \neq k \, \Longrightarrow a_k \neq a_j$.
\end{remark}
As mentioned in the beginning of the current section, a single rollout provides rather little information about an action particularly due to the combinatorial explosion in the branching factor of possible actions of the agent, randomness, unavailable information etcetera. Normally a large, yet comparable with total resource limitations, number of rollouts (or trials) is simulated (or attempted) to evaluate the actions at various states. The challenging question that the Geiringer-like theorem in \cite{MitavRowGeirNewMain} addresses is how one can take the full advantage of the available limited size sample of rollouts (or trials). In the language of evolutionary computing community such samples are known as \emph{populations}.
\begin{definition}\label{popOfRolloutsDefn}
Given a state in question $\vec{s} = (s, \vec{\alpha})$ and a sequence $\{\alpha_i\}_{i=1}^b$ of actions under evaluation (in the sense of definition~\ref{treeRootedByChanceNode}) then a \emph{population} $P$ with respect to the state $\vec{s} = (s, \vec{\alpha})$ and the sequence $\{\alpha_i\}_{i=1}^b$ is a sequence of rollouts $P = \{r_i^{l(i)}\}_{i=1}^b$ where $r_i = \{(\alpha_i, \, s_1^i, \, s_2^i, \ldots, s^i_{l(i)-1}, \, f_i)\}$. Just as in definition~\ref{RolloutDefn} we will assume that $s_k^i \neq s_q^j$ whenever $i \neq j$ (which, in accordance with definition~\ref{RolloutDefn}, is as strong as requiring that $s_k^i \neq s_q^j$ whenever $i \neq j$ or $k \neq q$)\footnote{The last assumption that all the states in a population are formally distinct (although they may be equivalent) will be convenient later to extend the crossover operators from pairs to the entire populations. This assumption does make sense from the intuitive point of view as well since the exact state in most situations involving randomness or incomplete information is simply unknown.} Moreover, we also assume that the terminal labels $f_i$ are also all distinct within the same population, i.e. for $i \neq j$ the terminal labels $f_i \neq f_j$\footnote{This assumption does not reduce any generality since one can choose an arbitrary (possibly a many to one) assignment function $\phi: \Sigma \rightarrow \mathbb{Q}$, yet the complexity of the statements of our main theorems will be mildly alleviated.} In a very special case when $s_j^i \sim s_k^q \Longrightarrow j=k$ we will say that the population $P$ is \emph{homologous}. Loosely speaking, a homologous population is one where equivalent states can not appear at different ``heights".
\end{definition}
\begin{remark}\label{popOfRolloutsRem}
Each rollout $r_i^{l(i)}$ in definition~\ref{popOfRolloutsDefn} is started with the corresponding move $\alpha_i$ of the sequence of moves under evaluation (see definition~\ref{treeRootedByChanceNode}). It is clear that if one were to permute the rollouts without changing the actual sequences of states the corresponding populations should provide identical values for the corresponding actions under evaluation. In fact, most authors in evolutionary computation theory (see \cite{VoseM}, for instance) do assume that such populations are equivalent and deal with the corresponding equivalence classes of multisets corresponding to the individuals (these are sequences of rollouts). Nonetheless, when dealing with finite-population Geiringer-like theorems it is convenient for technical reasons which will become clear when the proof is presented (see also \cite{MitavRowGeirMain} and \cite{MitavRowGeirGenProgr}) to assume the \emph{ordered multiset model} i.e. the populations are considered formally \emph{distinct} when the individuals are permuted. Incidentally, ordered multiset models are useful for other types of theoretical analysis in \cite{ShmittL1} and \cite{ShmittL2}.
\end{remark}
\begin{example}\label{PopRolloutEx}
A typical population with the convention as in remark~\ref{popOfRolloutsRem} might look as in figure~\ref{PopOfRollsFig}. The height of the leftmost rollout in figure~\ref{PopOfRollsFig} would then be $5$ since it contains $5$ states. The reader can easily see that the heights of the rollouts in this population read from left to right are $5$, $4$, $3$, $5$, $3$, $1$ and $4$ respectively.
\end{example}
The main idea is that equivalent states should be interchangeable due to the unavailable information, as discussed above. In the language of evolutionary computing, such a swap of states is called a crossover. In order to obtain the most out of a sample (population in our language) of the parallel rollouts it is desirable to explore all possible populations obtained by making various swaps of the corresponding rollouts at the equivalent positions. Computationally this task seems expensive if one were to run the type of genetic programming described precisely below, yet, it turns out that we can predict exactly what the limiting outcome of this ``mixing procedure" would be.\footnote{In the finite-population version of the Geiringer-like theorem we will also need to ``inflate" the population first and then take the limit of a sequence of these limiting procedures as the inflation factor increases. All of this will be presented below in sufficient detail.} We now continue with the rigorous definitions of crossover.
Representation of rollouts suggested in remark~\ref{crossoverConvRepresRem} is convenient to define crossover operators for two given rollouts. We will introduce two crossover operations below.
\begin{definition}\label{rolloutPartCrossDefn}
Given two rollouts $$r_1 = (\alpha_1, \, (i_1, \, a_1), \, (i_2, \, a_2), \ldots, (i_{t(1)-1}, a_{t(1)-1}), \, f)$$ and $$r_2 = (\alpha_2, \, (j_1, \, b_1), \, (j_2, \, b_2), \ldots, (j_{t(2)-1}, b_{t(2)-1}), \, g)$$ of lengths $t(1)$ and $t(2)$ respectively that share no state in common (i.e., as in definition~\ref{RolloutDefn}, ) there are two (non-homologous) crossover (or recombination) operators we introduce here. For an equivalence class label $m \in \mathbb{N}$ and letters $c, \, d \in A$ define the \emph{one-point non-homologous crossover} transformation $\chi_{m, \, c, \, d}(r_1, \, r_2) = (t_1, \, t_2)$ where \\$t_1 = (\alpha_1, \, (i_1, \, a_1), \ldots, (i_{k-1}, \, a_{k-1}), \, (j_q, \, b_q), \, (j_{q+1}, \, b_{q+1}), \ldots, (j_{t(2)-1}, b_{t(2)-1}), \, g)$ and\\ $t_2 = (\alpha_2, \, (j_1, \, b_1), \ldots, (j_{q-1}, \, b_{q-1}), \, (i_k, \, a_k), \, (i_{k+1}, \, a_{k+1}), \ldots, (i_{t(1)-1}, a_{t(1)-1}), \, f)$ if [$i_k = j_q = m$ and either $(a_k = c$ and $b_q = d)$ or $(a_k = d$ and $b_q = c)$] and $(t_1, \, t_2) = (r_1, \, r_2)$ otherwise.

Likewise, we introduce a \emph{single position swap crossover} $\nu_{m, \, c, \, d}(r_1, \, r_2) = (v_1, \, v_2)$ where\\ $v_1 = (\alpha_1, \, (i_1, \, a_1), \ldots, (i_{k-1}, \, a_{k-1}), \, (j_q, \, b_q), \, (i_{k+1}, \, a_{k+1}), \ldots, (i_{t(1)-1}, a_{t(1)-1}), \, f)$ \\while\\ $v_2 = (\alpha_2, \, (j_1, \, b_1), \ldots, (j_{q-1}, \, b_{q-1}), \, (i_k, \, a_k), \, (j_{q+1}, \, b_{q+1}), \ldots, (j_{t(2)-1}, b_{t(2)-1}), \, g)$ if [$i_k = j_q = m$ and either $(a_k = c$ and $b_q = d)$ or $(a_k = d$ and $b_q = c)$] and $(v_1, \, v_2) = (r_1, \, r_2)$ otherwise. In addition, a singe swap crossover is defined not only on the pairs of rollouts but also on a single rollout swapping equivalent states in the analogous manner: If $$r = (\alpha, \, (i_1, \, a_1), \, (i_2, \, a_2), \ldots, (i_{j-1}, \, a_{j-1}), \, (i_j, \, a_j), \, (i_{j+1}, \, a_{j+1}), \ldots $$$$ \ldots, (i_{k-1}, \, a_{k-1}), \, (i_k, \, a_k), \, (i_{k+1}, \, a_{k+1}), \ldots, (i_{t(1)-1}, a_{t(1)-1}), \, f)$$ and [$i_j = i_k$ and either $(a_j = c$ and $a_k = d)$ or $(a_j = d$ and $a_k = c)$] then $$\nu_{m, \, c, \, d}(r) = (\alpha, \, (i_1, \, a_1), \, (i_2, \, a_2), \ldots, (i_{j-1}, \, a_{j-1}), \, (i_j, \, a_k), \, (i_{j+1}, \, a_{j+1}), \ldots $$$$ \ldots, (i_{k-1}, \, a_{k-1}), \, (i_k, \, a_j), \, (i_{k+1}, \, a_{k+1}), \ldots, (i_{t(1)-1}, a_{t(1)-1}), \, f)$$ and, of course, $\nu_{m, \, c, \, d}(r)$ fixes $r$ (i.e. $\nu_{m, \, c, \, d}(r) = r$) otherwise.
\end{definition}
\begin{remark}\label{distConvRem}
Notice that definition~\ref{rolloutPartCrossDefn} makes sense thanks to the assumption that no rollout contains an identical pair of states in definition~\ref{RolloutDefn}.
\end{remark}
Just as in case of defining crossover operators for pairs of rollouts, thanks to the assumption that all the states in a population of rollouts are formally distinct (see definition~\ref{popOfRolloutsDefn}), it is easy to extend definition~\ref{rolloutPartCrossDefn} to the entire populations of rollouts.
Intuitively, the one-point crossover transformations correspond to the fact that due to the unavailable and unpredictable information, either of the alternative courses of events could take place following either of the indistinguishable states (i.e. equivalent states) while the single swap crossovers correspond to the fact that either of the non-observable components of the state could have arose. Thus, to get the most informative picture out of the sample of rollouts one would want to run the genetic programming routine without selection and mutation and using only the crossover operators specified above for as long as possible and then, in order to evaluate a certain action $\alpha$, collect the weighted average of the terminal values (i. e. the values assigned to the terminal labels via some real-valued assignment function) of all the rollouts starting with the action $\alpha$ that ever occurred in the process. We now describe precisely what the process is and give an example.
\begin{definition}\label{recombActOnPopsDef}
Given a population $P$ and a transformation of the form $\chi_{i, \, x, \, y}$, there exists at most one pair of distinct rollouts in the population $P$, namely the pair of rollouts $r_1$ and $r_2$ such that the state $(i, \, x)$ appears in $r_1$ and the state $(i, \, y)$ appears in $r_2$. If such a pair exists, then we define the recombination transformation $\chi_{i, \, x, \, y}(P) = P'$ where $P'$ is the population obtained from $P$ by replacing the pair of rollouts $(r_1, \, r_2)$ with the pair $\chi_{i, \, x, \, y}(r_1, \, r_2)$ as in definition~\ref{rolloutPartCrossDefn}. In any other case we do not make any change, i.e. $\chi_{i, \, x, \, y}(P) = P$. The transformation $\nu_{i, \, x, \, y}(P)$ is defined in an entirely analogous manner with one more amendment: if the states $(i, \, x)$ and $(i, \, y)$ appear within the same individual (rollout), call it $$r = (\alpha, \, (j_1, \, a_1), \, (j_2, \, a_2), \ldots, (i, \, x), \ldots, \, (i, \, y), \ldots, (i_{t(1)-1}, a_{t(1)-1}), \, f),$$ and the state $(i, \, x)$ precedes the state $(i, \, y)$, then these states are interchanged obtaining the new rollout $$r' = (\alpha, \, (j_1, \, a_1), \, (j_2, \, a_2), \ldots, (i, \, y), \ldots, \, (i, \, x), \ldots, (i_{t(1)-1}, a_{t(1)-1}), \, f).$$ Of course, it could be that the state $(i, \, y)$ precedes the state $(i, \, x)$ instead, in which case the definition would be analogous: if $$r = (\alpha, \, (j_1, \, a_1), \, (j_2, \, a_2), \ldots, (i, \, y), \ldots, \, (i, \, x), \ldots, (i_{t(1)-1}, a_{t(1)-1}), \, f)$$ then replace the rollout $r$ with the rollout $$r'=(\alpha, \, (j_1, \, a_1), \, (j_2, \, a_2), \ldots, (i, \, x), \ldots, \, (i, \, y), \ldots, (i_{t(1)-1}, a_{t(1)-1}), \, f).$$
\end{definition}
\begin{remark}\label{BijectRem}
It is very important for the general finite-population Geiringer theorem that each of the crossover transformations $\chi_{i, \, x, \, y}$ and $\nu_{i, \, x, \, y}$ is a bijection on their common domain, that is the set of all populations of rollouts (see \cite{MitavRowGeirMain} or \cite{MitavRowGeirNewMain} for details on the finite-population version of the theorem that's based on Markov chains induced by group actions). As a matter of fact, the reader can easily verify by direct computation from definitions~\ref{recombActOnPopsDef} and \ref{rolloutPartCrossDefn} that each of the transformations $\chi_{i, \, x, \, y}$ and $\nu_{i, \, x, \, y}$ is an involution on its domain, i.e. $\forall \, i, \, x, \, y$ we have $\chi_{i, \, x, \, y}^2 = \nu_{i, \, x, \, y}^2 = \mathbf{1}$ where $\mathbf{1}$ is the identity transformation.
\end{remark}
Examples below illustrate the important extension of recombination operators to arbitrary populations pictorially.
\begin{example}\label{popCrossEx1}
Continuing with example~\ref{PopRolloutEx}, suppose we were to apply the recombination (crossover) operator $\chi_{1, \, c, \, d}$ to the population of seven rollouts pictured in figure~\ref{PopOfRollsFig}. The unique location of states $(1, \, c)$ and $(1, \, d)$ in the population is emphasized by the boxes in figure~\ref{PopOfRollsFigSq1} below. After applying the crossover operator $\chi_{1, \, c, \, d}$ we obtain the population pictured on figure~\ref{PopOfRollsAfterCross1}.
\begin{figure} 
\centerline{\includegraphics[height=10cm]{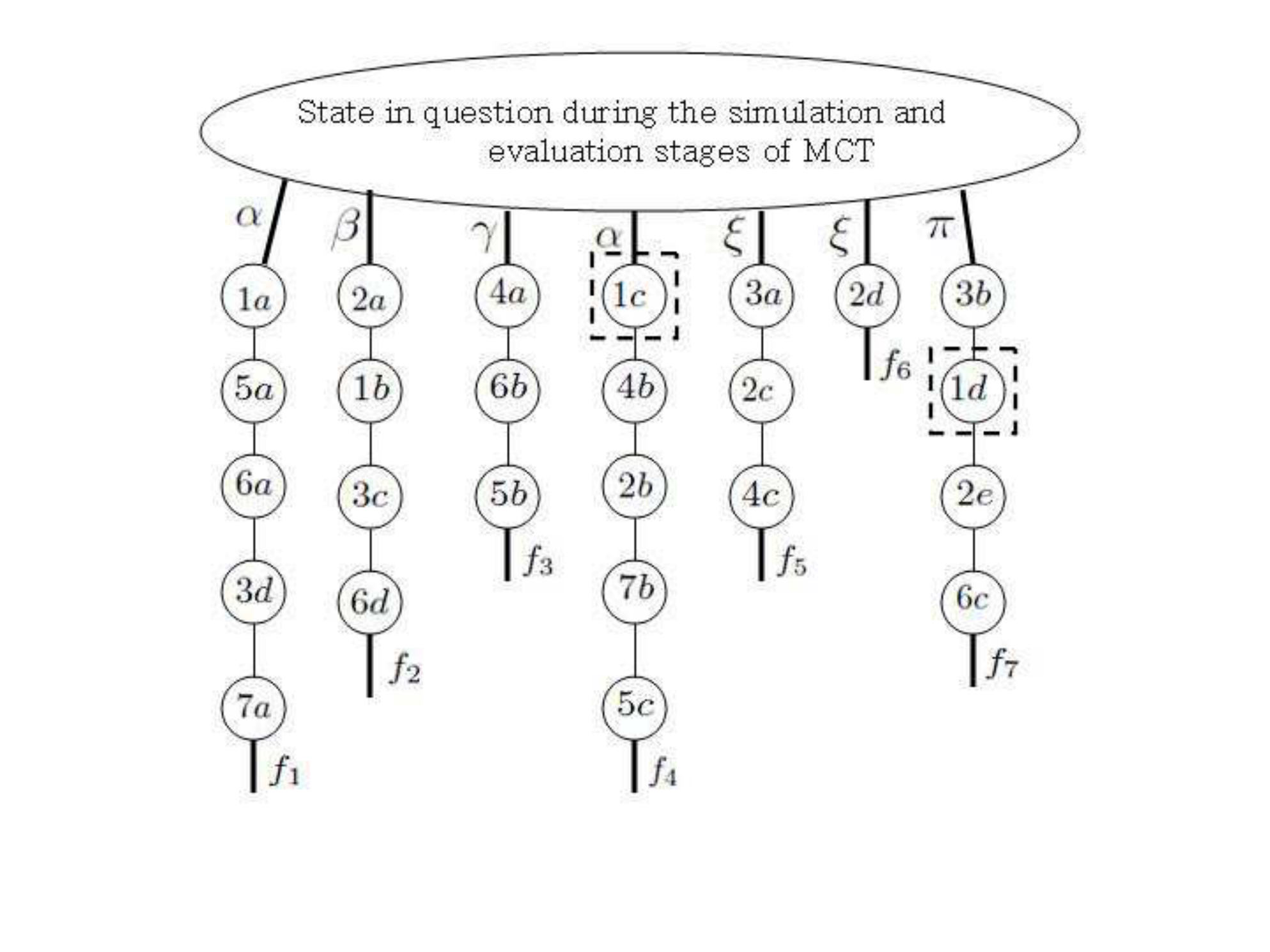}}
\caption{The unique states $(1, \, c)$ and $(1, \, d)$ in the population pictured in figure~\ref{PopOfRollsFig} are enclosed in dashed squares.}
\label{PopOfRollsFigSq1}
\end{figure}
\begin{figure}[htp]
\centerline{\includegraphics[height=10cm]{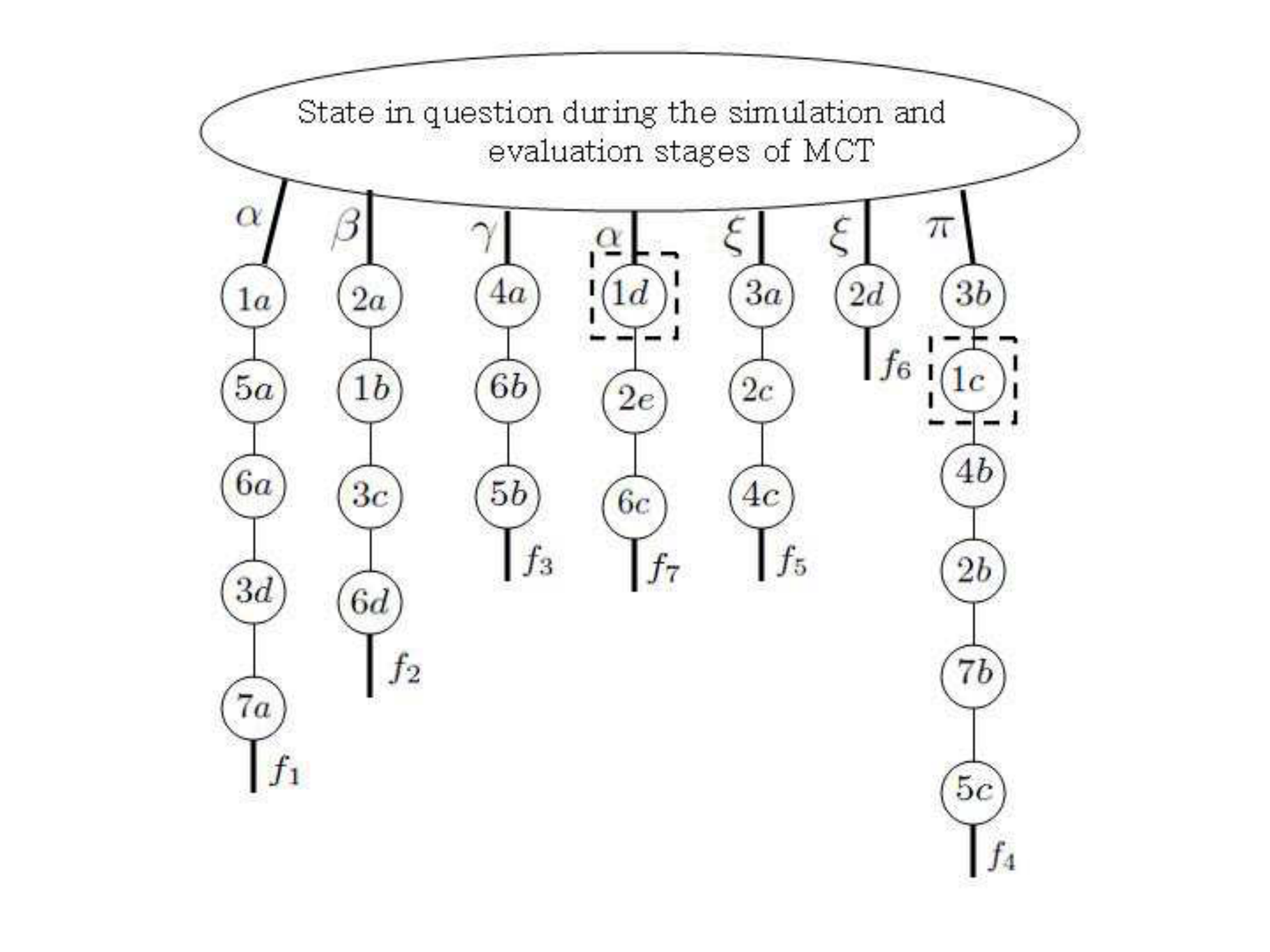}}
\caption{The subrollouts rooted at the states $(1, \, c)$ and $(1, \, d)$ in the population pictured in figure~\ref{PopOfRollsFigSq1} are pruned and then swapped.}
\label{PopOfRollsAfterCross1}
\end{figure}
On the other hand, applying the crossover transformation $\nu_{1, \, c, \, d}$ to the population in figures~\ref{PopOfRollsFig} and \ref{PopOfRollsFigSq1} results in the population pictured on figure~\ref{PopOfRollsAfterCross2}.
\begin{figure}[htp]
\centerline{\includegraphics[height=10cm]{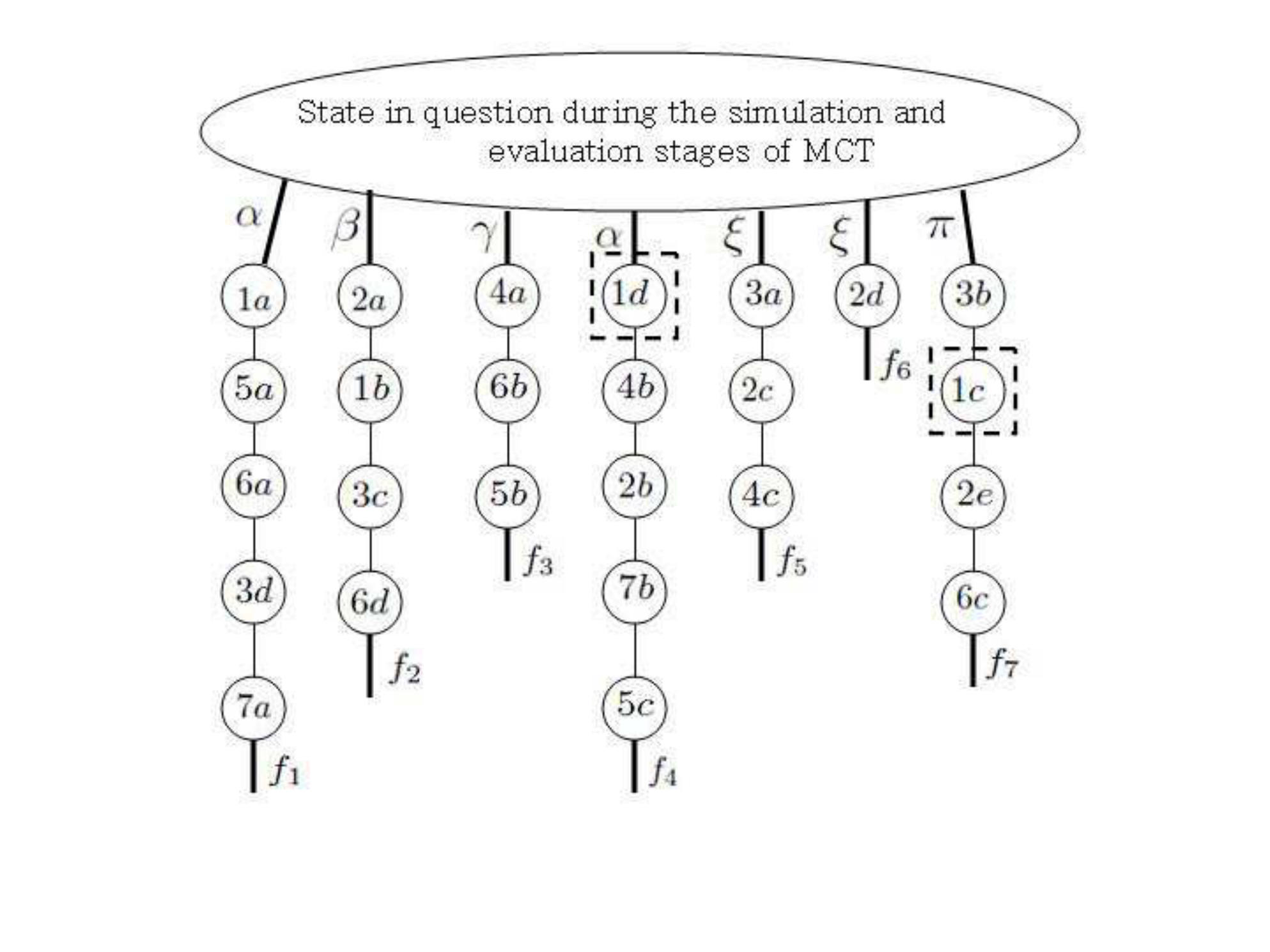}}
\caption{The uniquely positioned labels $(1, \, c)$ and $(1, \, d)$ are enclosed within the dashed squares in figure~\ref{PopOfRollsFigSq1} are swapped.}
\label{PopOfRollsAfterCross2}
\end{figure}
\end{example}
\begin{example}\label{popCrossEx2}
Consider now the population $Q$ pictured in figure~\ref{PopOfRollsFig2}. Suppose we apply the transformations $\chi_{6, \, a, \, b}$ and $\nu_{6, \, a, \, b}$ to the population $Q$.
\begin{figure}[htp]
\centerline{\includegraphics[height=10cm]{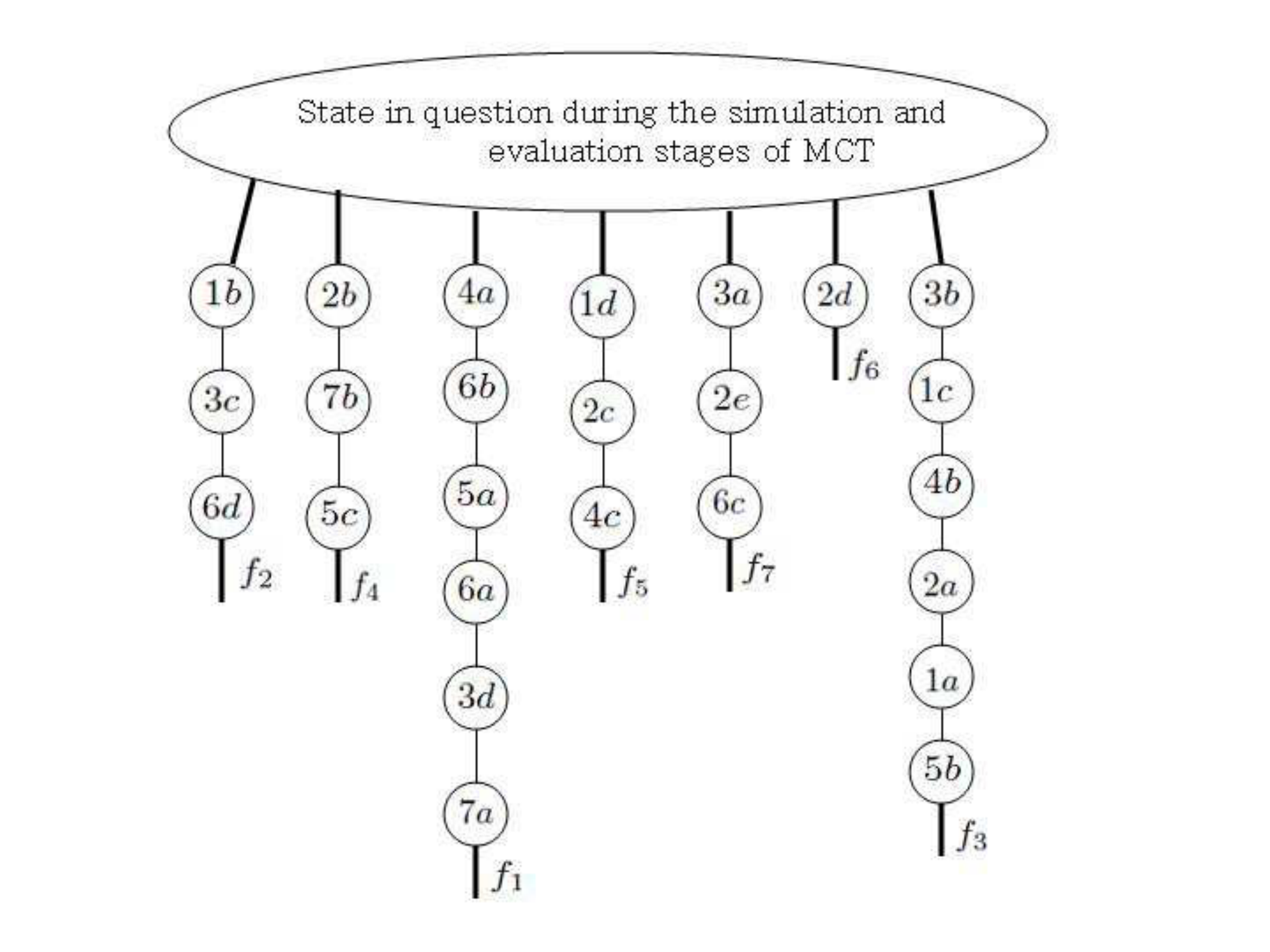}}
\caption{A population of rollouts $Q$.}
\label{PopOfRollsFig2}
\end{figure}
The states $(6, \, a)$ and $(6, \, b)$ are enclosed within the dashed squares in figure~\ref{PopOfRollsFig3}.
\begin{figure}[htp]
\centerline{\includegraphics[height=10cm]{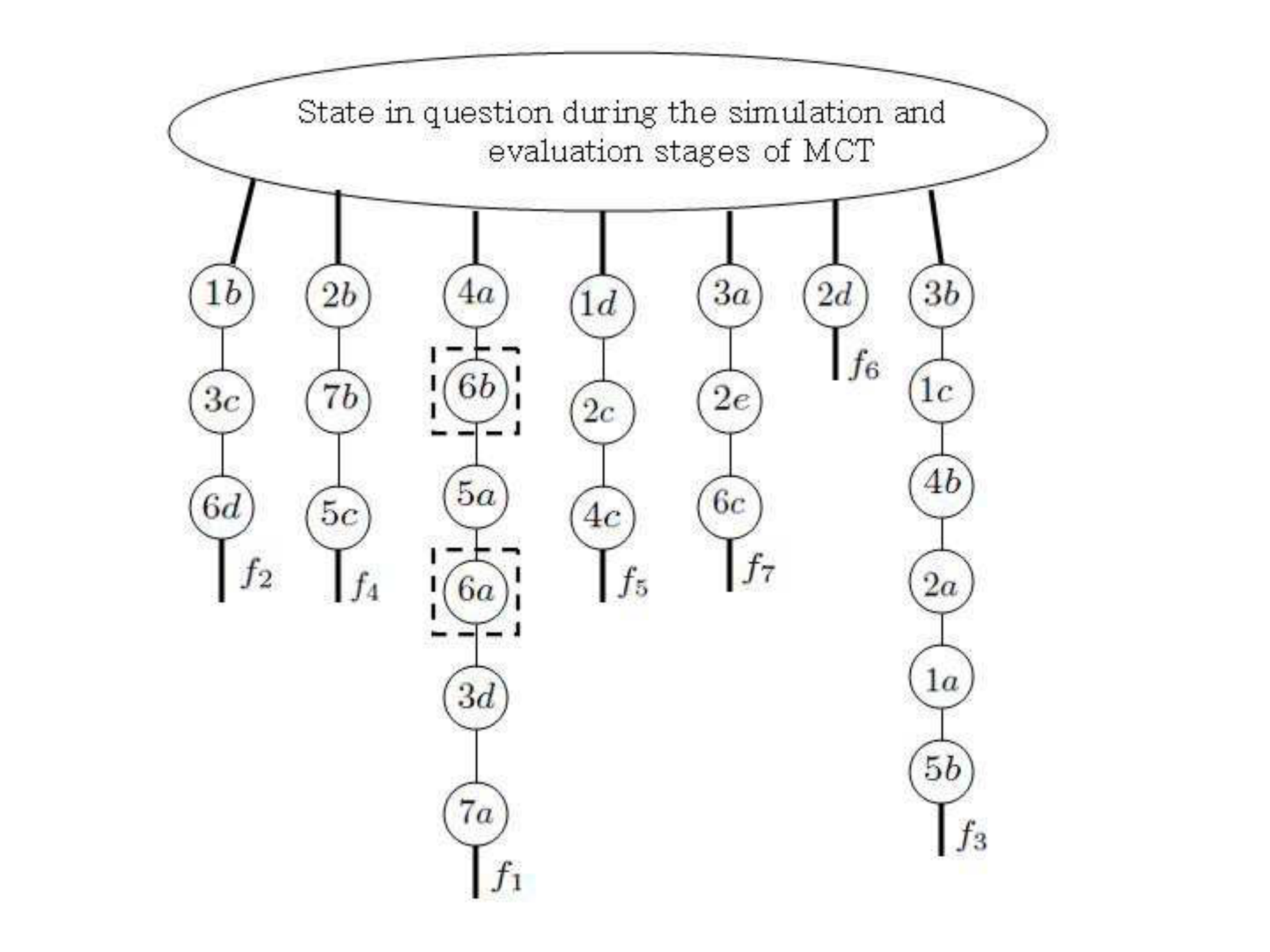}}
\caption{The uniquely positioned labels $(6, \, a)$ and $(6, \, b)$ are enclosed within the dashed squares.}
\label{PopOfRollsFig3}
\end{figure}
Since these states appear within the same rollout, according to definition~\ref{recombActOnPopsDef}, the crossover transformation $\chi_{6, \, a, \, b}$ fixes the population $Q$ (i.e. $\chi_{6, \, a, \, b}(Q)=Q$). On the other hand, the population $\nu_{6, \, a, \, b}(Q)$ is pictured on figure~\ref{PopOfRollsAfterCross3}.
\begin{figure}[htp]
\centerline{\includegraphics[height=10cm]{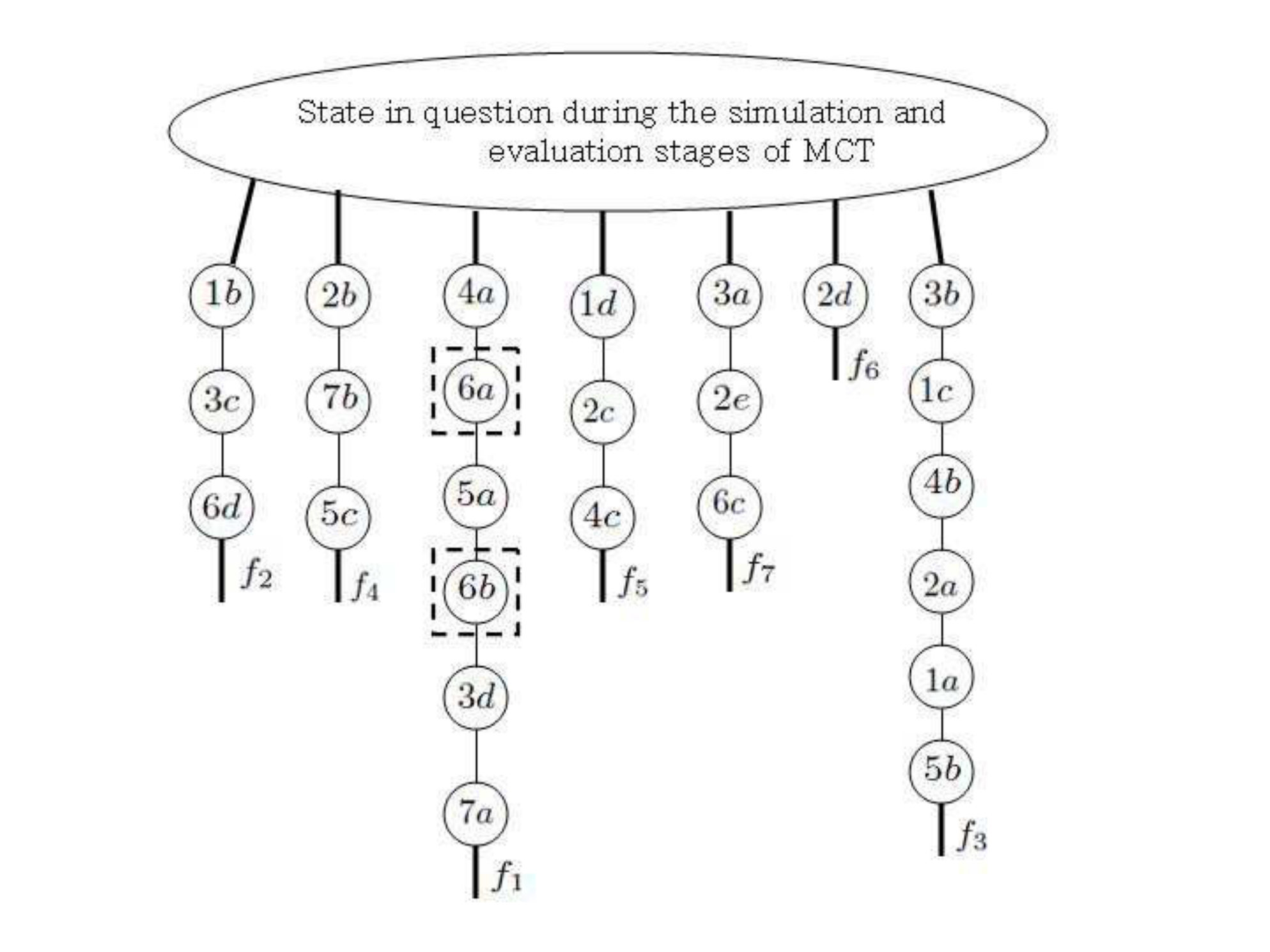}}
\caption{The uniquely positioned labels $(6, \, a)$ and $(6, \, b)$ which are enclosed within the dashed squares on figure~\ref{PopOfRollsFig3} are interchanged to obtain the population $\nu_{6, \, a, \, b}(Q)$ pictured above.}
\label{PopOfRollsAfterCross3}
\end{figure}
\end{example}
Evidently, running the ``genetic programming" routine with recombination only, as described above, for a very long time is computationally expensive, but, fortunately, thanks to to the Geiringer-like theorem in \cite{MitavRowGeirMain} we can predict the limiting frequency of occurrence of various ``schemata" or rollouts as the iteration time $t \rightarrow \infty$. Furthermore, this prediction leads to dynamic parallel action-evaluation algorithms based on an exponentially larger sample of rollouts obtained after such an ``infinitely long time" GP-routine, that resemble Hebbian learning mechanisms as we intend to demonstrate through Andree Ehresmann's ``Memory Evolutive Systems" model, the essentials of which will be briefly described in the next section. We now proceed with the relevant definitions.
\begin{definition}\label{schemaForMCTPopDefHolland}
Given a state $(s, \vec{\alpha})$ in question (see definition~\ref{treeRootedByChanceNode}), a rollout \emph{Holland-Poli schema} is a sequence consisting of entries from the set$\vec{\alpha} \cup \mathbb{N} \cup \{\#\} \cup \Sigma$ of the form $h = \{x_i\}_{i=1}^k$ for some $k \in \mathbb{N}$ such that for $k>1$ we have $x_1 \in \vec{\alpha}$, $x_i \in \mathbb{N}$ when $1 < i < k$ represents an equivalence class of states, and $x_k \in \{\#\} \cup \Sigma$ could represent either a terminal label if it is a member of the set of terminal labels $\Sigma$, or any substring defining a valid rollout if it is a $\#$ sign.\footnote{This notion of a schema is somewhat of a mixture between Holland's and Poli's notions.} For $k=1$ there is a unique schema of the form $\#$. Every schema uniquely determines a set of rollouts
$$S_h = \begin{cases}
\{(x_1, \, (x_2, a_2), \, (x_3, a_3), \ldots, (x_{k-1}, a_{k-1}), x_k) \, \\| \, a_i \in A \text{ for } 1<i<k\} & \text{ if } k>1 \text{ and } x_k \in \Sigma\\
\{(x_1, \, (x_2, a_2), \, (x_3, a_3), \ldots, (x_{k-1}, a_{k-1}), \, \\(y_k, a_k), \, (y_{k+1}, a_{k+1}), \ldots, f)\\
\, | \, a_i \in A \text{ for } 1<i<k, \, y_j \in \mathbb{N} \text{ and } a_j \in A\} & \text{ if } k > 1 \text{ and } x_k = \#\\
\text{the entire set of all possible rollouts} & \text{ if } k = 1 \text{ or, equivalently, } h=\#.
\end{cases}$$ that fit the schema in the sense mentioned above. We will often abuse the language and use the same word schema to mean either the schema $h$ as a formal sequence as above or schema as a set $S_h$ of rollouts which fit the schema. For example, if $h$ and $h^*$ is a schema, we will write $h \cap h^*$ as a shorthand notation for $S_h \cap S_{h^*}$ where $\cap$ denotes the usual intersection of sets. Just as in definition~\ref{RolloutDefn}, we will say that $k-1$, the number of states in the schema $h$ is the \emph{height} of the schema $h$.
\end{definition}
We illustrate the important notion of a schema with an example below:
\begin{example}\label{schemaDefEx}
Suppose we are given a schema $h=(\alpha, \, 1, \, 2, \, \#)$. Then the rollouts\\ $(\alpha, \, 1a, \, 2c, \, 5a, \, 3c, \, f)$ and $(\alpha, \, 1d, \, 2a, \, 3a, \, 3d, \, g) \in S_h$ or one could say that both of them fit the schema $h$. On the other hand the rollout $(\beta, \, 1a, \, 2c, \, 5a, \, 3c, \, f) \notin S_h$ (or does not fit the schema $h$) unless $\alpha = \beta$. A rollout $(\alpha, \, 1a, \, 3a, \, 5a, \, 3c, \, f) \notin S_h$ does not fit the schema $h$ either since $x_2 = 2 \neq 3$. Neither of the rollouts above fit the schema $h^* = (\alpha, \, 1, \, 2, \, f)$ since the appropriate terminal label is not reached in the $4^{\text{th}}$ position. An instance of a rollout which fits the schema $h^*$ would be $(\alpha, \, 1c, \, 2b, \, f)$.
\end{example}
In evolutionary computation Geiringer-like results address the limiting frequency of occurrence of a set of individuals fitting a certain schema after repeated applications of crossover transformations as in definition~\ref{recombActOnPopsDef}. (see \cite{PoliGeir}, \cite{MitavRowGeirMain}, \cite{MitavRowGeirGenProgr} and \cite{MitavRowGeirNewMain}). The finite population Geiringer theorem, originated in \cite{MitavRowGeirMain} and extended to non-homogenous time Markov processes in section 6 of \cite{MitavRowGeirNewMain}, states that the unique stationary distribution of the Markov chain of all populations potentially encountered in the process of repeated crossover transformations is the uniform distribution and holds under rather mild assumptions on the family of all such recombination transformations. We invite the readers to study \cite{MitavRowGeirNewMain} for an in-depth understanding. In the current paper we present only a very brief description of the corresponding Markov chain and the notion of the ``limiting frequency of occurrence" that is sufficient for a surface-level understanding of what the theorem says. The Geiringer-like theorem that motivates the novel parallel algorithms resembling Hebbian learning mechanisms is stated purely in terms of the quantities appearing in the following few definitions.
\begin{definition}\label{popStateCountDownDef}
For any action under evaluation $\alpha$ define a set-valued function $\alpha \downarrow$ from the set $\Omega^b$ of populations of rollouts to the power set of the set of natural numbers $\mathcal{P}(\mathbb{N})$ as follows: $\alpha \downarrow(P) = \{i \, | \, i \in \mathbb{N}$ and at least one of the rollouts in the population $P$ fits the Holland schema $(\alpha, \, i, \, \#)\}$. Likewise, for an equivalence class label $i \in \mathbb{N}$ define a set valued function on the populations of size $b$, as $i \downarrow (P) = \{j \, | \, \exists \, x$ and $y \in A$ and a rollout $r$ in the population $P$ such that $r = (\ldots, (i, \, x), \, (j, \, y), \ldots) \, \} \cup \{f \, | \, f \in \Sigma$ and $\exists$ an $x \in A$ and a rollout $r$ in the population $P$ such that $r = (\ldots, (i, \, x), \, f) \, \}$. In words, the set $i \downarrow (P)$ is the set of all equivalence classes together with the terminal labels which appear after the equivalence class $i$ in at least one of the rollouts from the population $P$. Finally, introduce one more function, namely $i \downarrow_\Sigma: \Omega^b \rightarrow \mathbb{N} \cup \{0\}$ by letting $i \downarrow_\Sigma(P) = |\{f \, | \, f \in \Sigma \cap i \downarrow (P)\}|$, that is, the total number of terminal labels (which are assumed to be all formally distinct for convenience) following the equivalence class $i$ in a rollout of the population $P$.
\end{definition}
We illustrate definition~\ref{popStateCountDownDef} in example~\ref{popSchemaDownFunctEx} below.
\begin{example}\label{popSchemaDownFunctEx}
Continuing with example~\ref{PopRolloutEx}, we return to the population $P$ in figure~\ref{PopOfRollsFig}. From the picture we see that the only equivalence class $i$ such that a rollout from the population $P$ fits the Holland schema $(\alpha, \, i, \, \#)$ is $i = 1$ so that $\alpha \downarrow (P) = \{1\}$. Likewise, the only equivalence class following the action $\beta$ is $2$, the only equivalence class following the action $\gamma$ is $4$ and the only one following $\pi$ is $3$ so that $\beta \downarrow (P) = \{2\}$, $\gamma \downarrow (P) = \{4\}$ and $\pi \downarrow (P) = \{3\}$. The only equivalence classes $i$ following the action $\xi$ in the population $P$ are $i=3$ and $i=2$ so that the set $\xi \downarrow (P) = \{2, \, 3\}$.

Likewise the fragment $(1, \, a), (5, \, a)$ appears in the first (leftmost) rollout in $P$, $(1, \, b), (3, \, c)$ in the second rollout, $(1, \, c), (4, \, b)$ in the forth tollout and $(1, \, d), (2, \, e)$ in the last, seventh rollout. No other equivalence class or a terminal label follows the equivalence class of the state $1$ in the population $P$ and so it follows that $1 \downarrow (P) = \{5, \, 3, \, 4, \, 2\}$ and $1 \downarrow_\Sigma (P) = |\{\emptyset\}|=0$. Likewise, equivalence class $1$ follows the equivalence class $2$ in the second rollout, $7$ follows $2$ in the forth rollout, $4$ follows $2$ in the fifth rollout and $6$ follows $2$ in the last, seventh rollout. The only terminal label that follows the equivalence class $2$ is $f_6$ in the $6^{\text{th}}$ rollout. Thus we have $2 \downarrow (P) = \{1, \, 7, \, 4, \, 6, \, f_6\}$ and $2 \downarrow_\Sigma (P) = |\{f_6\}| = 1$. We leave the reader to verify that $$3 \downarrow (P) = \{7, \, 6, \, 2, \, 1\} \text{ so that } 3 \downarrow_\Sigma (P) = 0,$$ $$4 \downarrow (P) = \{6, \, 2, \, f_5\} \text{ so that } 4 \downarrow_\Sigma (P)  = 1,$$ $$5 \downarrow (P) = \{6, \, f_3, \, f_4\} \text{ and so } 5 \downarrow_\Sigma (P) = 2,$$ $$6 \downarrow (P) = \{3, \, 5, \, f_2, \, f_7\} \text{ and so } 6 \downarrow_\Sigma (P) = 2$$ and, finally, $7 \downarrow (P) = \{5, \, f_1\} \text{ so that } 7 \downarrow_\Sigma (P)  = 1$.
\end{example}
\begin{remark}\label{totalNumberOfTermLblsRem}
Note that according to the assumption that all the terminal labels within the same population are distinct (see definition~\ref{popOfRolloutsDefn} together with the comment in the footnote there). But then, since every rollout ends with a terminal label, we must have $\sum_{i=1}^{\infty}i \downarrow_\Sigma (P) = b$ (of course, only finitely many summands, namely these equivalence classes that appear in the population $P$ may contribute nonzero values to $\sum_{i=1}^{\infty}i \downarrow_\Sigma (P)$) where $b$ is the number of rollouts in the population $P$, i.e. the size of the population $P$. For instance, in example~\ref{popSchemaDownFunctEx} $b = 7$ and there are totally $7$ equivalence classes, namely $1, \, 2, \, 3, \, 4, \, 5, \, 6$ and $7$ that occur within the population in figure~\ref{PopOfRollsFig} so that we have $\sum_{i=1}^{\infty}i \downarrow_\Sigma (P) = \sum_{i=1}^7 i \downarrow_\Sigma (P) = 0 + 1 + 0 + 1 + 2 + 2 + 1 = 7 = b$.
\end{remark}
Another important and related definition we need to introduce is the following:
\begin{definition}\label{PopStateOrderSeqDef}
Given a population $P$ and integers $i$ and  $j \in \mathbb{N}$ representing equivalence classes, let $$\text{Order}(i \downarrow j, \, P) = \begin{cases}
0 & \text{if } i(P) = 0 \text{ or } j \notin i \downarrow (P) \\
|\{((i, a), \, (j, \, b)) \, | \, \text{ the segment } & \text{ }\\
((i, a), \, (j, \, b)) \text{ appears in one of the} & \text{ }\\
\text{rollouts in the population }P\}| & \text{ otherwise }
\end{cases}.$$
Loosely speaking, $\text{Order}(i \downarrow j, \, P)$ is the total number of times the equivalence class $j$ follows the equivalence class $i$ within the
population of rollouts $P$.

Likewise, given a population of rollouts $P$, an action $\alpha$ under evaluation and an integer $j \in \mathbb{N}$, let $$\text{Order}(\alpha \downarrow j, \, P) = \begin{cases}
0 & \text{if } i(P) = 0 \text{ or } j \notin \alpha \downarrow j \\
|\{(\alpha, \, (j, \, b)) \, | \, \text{ the segment } & \text{ }\\
(\alpha, \, (j, \, b)) \text{ appears in one of the} & \text{ }\\
\text{rollouts in the population }P\}| & \text{ otherwise }
\end{cases}.$$
Alternatively, $\text{Order}(\alpha \downarrow j, \, P)$ is the number of rollouts in the population $P$ fitting the rollout Holland schema $(\alpha, \, j, \, \#)$.
\end{definition}
We now provide an example to illustrate definition~\ref{PopStateOrderSeqDef}.
\begin{example}\label{PopStateOrderEx}
Continuing with example~\ref{popSchemaDownFunctEx} and population $P$ appearing in figure~\ref{PopOfRollsFig}, we recall that $\alpha \downarrow (P) = \{1\}$. we immediately deduce that $\text{Order}(\alpha, \, j, \, \#) = 0$ unless $j=1$. There are two rollouts, namely the first and the forth, that fit the schema $(\alpha, \, 1, \, \#)$ so that $\text{Order}(\alpha \downarrow 1, \, P) = 2$. Likewise, $\beta \downarrow (P) = \{2\}$ and exactly one rollout, namely the second one, fits the Holland schema $(\beta, \, 2, \, \#)$ so that $\text{Order}(\beta, \, j, \, \#) = 0$ unless $j=2$ while $\text{Order}(\beta \downarrow 2, \, P) = 1$. Continuing in this manner (the reader may want to look back at example~\ref{popSchemaDownFunctEx}), we list all the nonzero values of the function $\text{Order}(\text{action}, \Box, \, P)$ for the population $P$ in figure~\ref{PopOfRollsFig}: $\text{Order}(\gamma \downarrow 4, \, P) = \text{Order}(\xi \downarrow 3, \, P) = \text{Order}(\xi \downarrow 2, \, P) = \text{Order}(\pi \downarrow 3, \, P) = 1$.

Likewise, recall from example~\ref{popSchemaDownFunctEx}, that $1 \downarrow (P) = \{5, \, 3, \, 4, \, 2\}$ so that $\text{Order}(1 \downarrow j, \, P) = 0$ unless $j = 5$ or $j=3$ or $j=4$ or $j=1$. It happens so that a unique rollout exists in the population $P$ fitting each fragment $(1, \, (j, \, \text{something in }A))$ for $j = 5$, $j=3$, $j=4$ and $j=2$ respectively, namely the first, the second, the forth and the last (seventh) rollouts. According to definition~\ref{PopStateOrderSeqDef}, we then have $\text{Order}(1 \downarrow 5, \, P) = \text{Order}(1 \downarrow 3, \, P) = \text{Order}(1 \downarrow 4, \, P) = \text{Order}(1 \downarrow 2, \, P) = 1$. Analogously, $2 \downarrow (P) = \{1, \, 7, \, 4, \, 6, \, f_6\}$ so that $\text{Order}(2 \downarrow j, \, P) = 0$ unless $j = 1, \, 7, \, 4$ or $6$. The only rollout in the population $P$ involving the fragment with $1$ following $2$ is the second one, the only one involving $7$ following $2$ is the forth, the only one involving $4$ following $2$ is the fifth, and the only one involving $6$ following $2$ is the last (the seventh) rollouts respectively so that $\text{Order}(2 \downarrow 1, \, P) = \text{Order}(2 \downarrow 7, \, P) = \text{Order}(2 \downarrow 4, \, P) = \text{Order}(2 \downarrow 6, \, P) = 1$. Continuing in this manner, we list all the remaining nonzero values of the ``Order" function introduced in definition~\ref{PopStateOrderSeqDef} for the population $P$ in figure~\ref{PopOfRollsFig}: $$\text{Order}(3 \downarrow 7, \, P) = \text{Order}(3 \downarrow 6, \, P) = \text{Order}(3 \downarrow 2, \, P) = \text{Order}(3 \downarrow 1, \, P) = 1,$$
$$\text{Order}(4 \downarrow 6, \, P) = \text{Order}(4 \downarrow 2, \, P) = 1,$$
$$\text{Order}(5 \downarrow 6, \, P) = \text{Order}(6 \downarrow 3, \, P) = \text{Order}(6 \downarrow 5, \, P) = \text{Order}(7 \downarrow 5, \, P) =1.$$
\end{example}
\begin{remark}\label{equivClassIndepRem}
It must be noted that all the functions introduced in definitions~\ref{popStateCountDownDef} and \ref{PopStateOrderSeqDef} remain invariant if one were to apply the ``primitive" recombination transformations as in definition~\ref{recombActOnPopsDef} to the population in the argument. More explicitly, given any population of rollouts $Q$ that's obtained from the initial population $P$ after repeated application of recombination transformations in definition \ref{recombActOnPopsDef}, an action $\alpha$ under evaluation, an equivalence class $i \in \mathbb{N}$, a Holland-Poli schema $h = (\alpha, \, i_1, \, i_2, \ldots, i_{k-1}, x_k)$ an integer $j$ with $1 \leq j \leq k$, we have $$\alpha \downarrow (P) = \alpha \downarrow (Q), \; i \downarrow (P) = i \downarrow (Q),$$$$i \downarrow_\Sigma (P) = i \downarrow_\Sigma (Q), \; \text{Order}(q \downarrow r, \, P) = \text{Order}(q \downarrow r, \, Q).$$
Indeed, the reader may easily verify that performing a swap of the elements of the same equivalence class, or of the corresponding subtrees pruned at equivalent labels, preserves all the states which are present within the population and creates no new ones. Moreover, the equivalence class sequel is also preserved and hence the invariance of the functions $\alpha \downarrow$ and $i \downarrow$ etc. follows.
\end{remark} 

We now briefly describe what the ``limiting frequency" of a schema of rollouts is as follows: Suppose we put a probability distribution, call it $\mu$, on the the set of all possible compositions of crossover transformations in definition~\ref{recombActOnPopsDef} such that identity transformation (i.e. the transformation that does not do anything, for instance, the composition of a one point crossover with itself) is chosen with a positive (it can be very tiny) probability. Now start with a population $P^0$ of rollouts, select a composition of recombination transformation $\theta$ with probability $\mu$, apply it to the population $P$ and obtain a new population $P^1 = \theta(P)$. Again select another transformation $\theta_1$ with probability $\mu$, apply it to the population $P^1$, obtain a new population $P^2 = \theta_1(P^1) = \theta_1 \circ \theta(P)$. Continue in this manner so that the new population at time $t$, $P^t = \theta_{t-1} \circ \theta_{t-2} \circ \ldots \circ \theta_1 \circ \theta(P^0)$.\footnote{It has been shown in section 6 of \cite{MitavRowGeirNewMain} that one can actually select an entire collection of such probability distributions on the family of compositions of recombination transformations and apply different distributions depending on the time as well as the history of populations up to the current time population (it must not depend on the current time population, of course).} Given a Holland-Poli schema $h$ of rollouts, for every population of rollouts $Q$ let $\mathcal{X}(h, \, Q)$ denote the total number of rollouts in the population $Q$ that fit the schema $h$. Since every one of the populations $P^t$ contains $b = |P^0|$ rollouts, the total number of rollouts encountered up to time $t$ is $bt$. The total number of rollouts fitting the Holland-Poli rollout schema $h$ up to the time $t$ is $\sum_{i=0}^{t}\mathcal{X}(h, \, P^t)$. We define the frequency of occurrence of the schema $h$ up to time $t$ as $\Phi(P, \, \{P^i\}_{i=1}^t, \, h) = \frac{\sum_{i=0}^{t}\mathcal{X}(h, \, P^t)}{bt}$ and the limiting frequency of occurrence of the Holland-Poli schema $h$ to be $\lim_{t \rightarrow \infty} \Phi(P^0, \, \{P^i\}_{i=1}^t, \, h).$ According to the general Geiringer theorem of \cite{MitavRowGeirMain} (see also the mildly extended version in \cite{MitavRowGeirNewMain} as in the previous footnote), the limiting frequency of occurrence of any schema (and, in fact, any specified subset) of rollouts always exists and is independent of the specific stochastic sample run with probability $1$. Computing this limit seems to be a very difficult (if not impossible) task for any given initial population $P^0$, however, if we ``inflate" the population $P^0$ by a factor of $m$ (i.e $m$-plicate every rollout in the population $P^0$: more explicitely, construct the new population $P$ by extending the alphabet by a factor of $m$ and for every rollout $x_i$ adding $m-1$ copies of the rollout $x_i$ to the population $P^0$, thereby obtaining a new population $P^0_m$)\footnote{Notice that inflating the population $P^0$ by a factor of $m \geq 1$ preserves all of the stochastic information within the population of samples: it only ``emphasizes" this information by a factor of $m$.} and then take the limit of the limiting frequencies of occurrence $\lim_{m \rightarrow \infty} \left(\lim_{t \rightarrow \infty}\Phi(P^0_m, \, \{P^i_m\}_{i=1}^t, \, h)\right)$ as the inflation factor $m \rightarrow \infty$, then the limit can be computed purely in terms of the quantities in definitions~\ref{popStateCountDownDef} and \ref{PopStateOrderSeqDef} as follows: if $h = (\alpha, \, i_1, \, i_2, \ldots, i_{k-1}, x_k)$, where $x_k \in \{\#\} \cup \Sigma$ is a given Holland-Poli schema then
\begin{equation}\label{GeiringerThmMainEq}
\lim_{m \rightarrow \infty}\lim_{t \rightarrow \infty}\Phi(P^0_m, \, \{P^i_m\}_{i=1}^t, \, h) = \frac{\emph{Order}(\alpha \downarrow i_1, \, P^0)}{b} \times$$$$ \times \left(\prod_{q = 2}^{k-1}\frac{\emph{Order}(i_{q-1} \downarrow i_q, \, P^0)}{\sum_{j \in i_{q-1} \downarrow} \emph{Order}(i_{q-1} \downarrow j, \, P^0) + i_{q-1} \downarrow_\Sigma (P^0)}\right) \cdot \text{\emph{LF}}(P^0, h)
\end{equation}
with probability $1$, where $$\text{\emph{LF}}(P^0, h) = \begin{cases}
1 & \text{if } x_k = \#\\
0 & \text{if } x_k = f \in \Sigma \text{ and } f \notin x_{k-1} \downarrow (P^0)\\
\emph{Fraction} & \text{if } x_k = f \in \Sigma \text{ and } f \in x_{k-1} \downarrow_{\Sigma} (P^0)
\end{cases}$$ where $$\emph{Fraction} = \frac{1}{\sum_{j \in i_{k-1} \downarrow (P^0)}\emph{Order}(i_{k-1} \downarrow j, \, P^0) +  i_{k-1} \downarrow_\Sigma (P^0)}$$ (we write ``LF" as short for ``Last Factor"). Furthermore, in the special case when the initial population $P^0$ is homologous (see definition~\ref{popOfRolloutsDefn}), one does not need to take the limit as $m \rightarrow \infty$ in the sense that $\lim_{t \rightarrow \infty}\Phi(P^0_m, \, h, \, t)$ is a constant independent of $m$ and its value is given by the right hand side of equation~\ref{GeiringerThmMainEq}.

An important comment is in order here: it is possible that the denominator of one of the fractions involved in the product is $0$. However, in such a case, the numerator is also $0$ and we adopt the convention (in equation~\ref{GeiringerThmMainEq}) that if the numerator is $0$ then, regardless of the value of the denominator (i.e. even if the denominator is $0$), then the fraction is $0$. As a matter of fact, a denominator of some fraction involved is $0$ if and only if one of the following holds: $\alpha(P) = 0$ or if there exists an index $q$ with $1 \leq q \leq k-1$ such that no state in the equivalence class of $i_q$ appears in the population $P$ (and hence in either of the inflated populations $P_m$).

\begin{figure}[htp]
\centerline{\includegraphics[height=10cm]{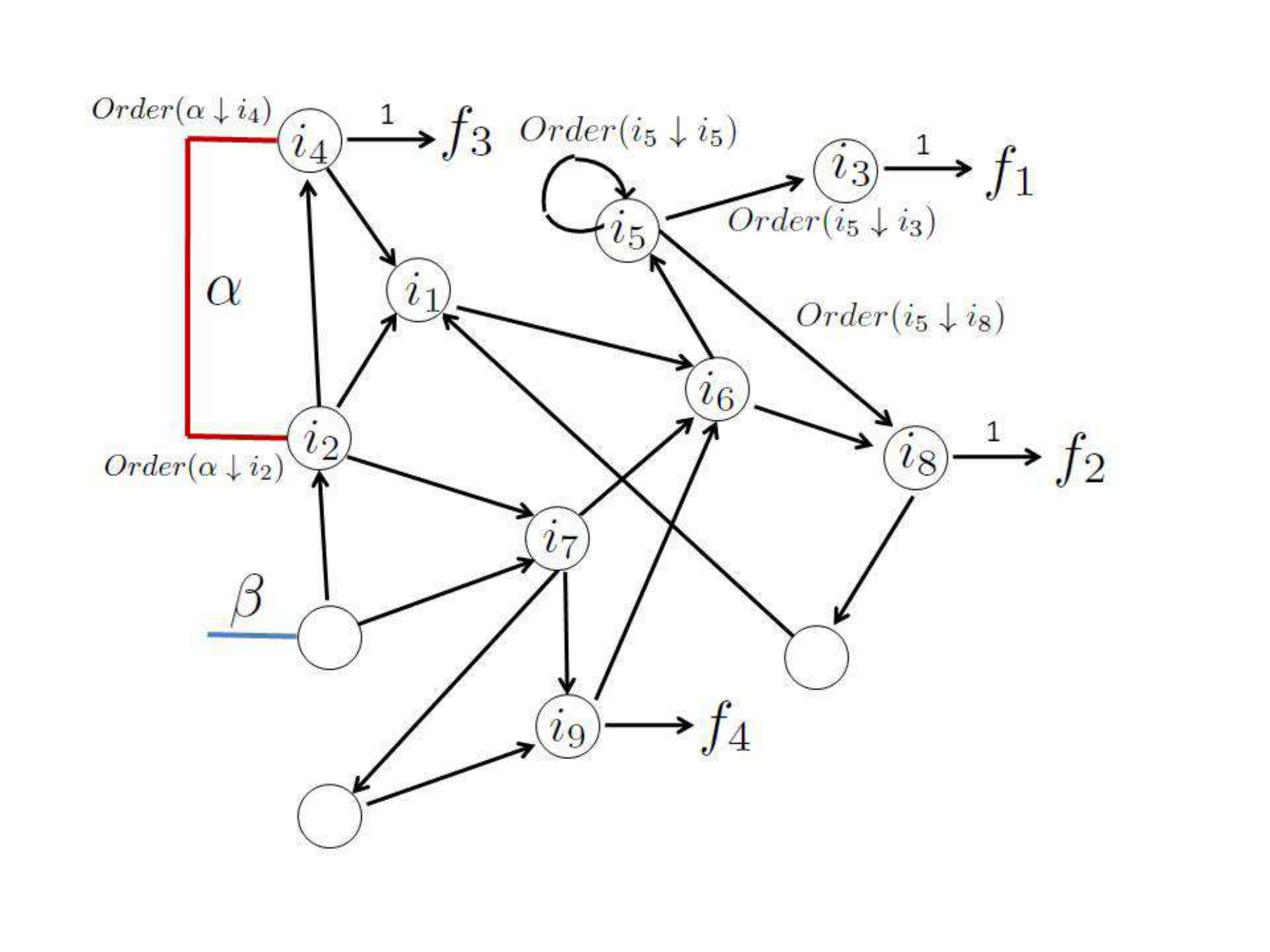}}
\caption{$\alpha$ and $\beta$ denote the actions under evaluation while $i_1, \, i_2, \ldots \in \mathbb{N}$ are the equivalence classes of states.}
\label{IllustrateAlgorithmPic}
\end{figure}

The Geiringer-like theorem above naturally motivates the following dynamic parallel algorithm for action evaluation after a number of simulations has been completed. Just as in Monte-Carlo tree search, as the learning agent keeps simulating rollouts (trials or random self-plays), it stores and dynamically updates a weighted labeled directed graph where the nodes are the similarity classes while a directed edge from a similarity class $i$ to the similarity class $j$ is added when some state from the similarity class $j$ follows a state from the similarity class $i$ within one of the simulated rollout. After being added the edge has weight $1$. Whenever a state with observable information $j$ follows a state with observable information $i$ again, the edge weight from $i$ to $j$ is incremented by $1$. Loops (or edges from a similarity class $i$ to itself) are allowed and their weights are incremented according to exactly the same rule with $i = j$. The same exactly rule applies to the terminal labels following a given edge. Either concurrently or periodically after a certain number of simulations (trials) independent agents, let's call them bugs, traverse the dynamically constructed digraph starting at various actions under evaluation and traveling through the digraph according to the following simple rule: if an agent is at the state labeled by $i$, the agent travels to a state labeled by $j \in i \downarrow (P)$ with the probability proportional to the directed edge weight from $i$ to $j$ that is, by construction, $\emph{Order}(i \downarrow j, \, P)$ where $P$ is the current sampled population. Of course, the situation is analogous if the bug starts at the actions under evaluation or travels to the terminal states (where it receives payoffs). The bug finishes its journey at terminal states and then updates the values $Q(s, \, \alpha)$ of the actions under evaluation that it has started traveling at according to the following simple rule: $Q(s, \, \alpha) := \frac{n(s, \, \alpha)}{n(s, \, \alpha)+1}Q(s, \, \alpha) + \frac{f_{term}}{n(s, \, \alpha)+1}$ where $n(s, \, \alpha)$ is the total number of times that the bugs updated the action value $Q(s, \, \alpha)$ up to the current, $n(s, \, \alpha)+1$st, bug and $f_{term}$ is the numerical value of the payoff where the current, $n(s, \, \alpha)+1$st, bug has finished its trip. the value $Q(s, \, \alpha)$ has been updated by the $n(s, \, \alpha)+1$st bug that started (or traveled) through that action under evaluation, increment $n(s, \, \alpha)$ by $1$, i.e. update $n(s, \, \alpha):=n(s, \, \alpha)+1$. It is not hard to see from equation~\ref{GeiringerThmMainEq} and the law of large numbers that as the number of bugs updating the payoff values of the action $\alpha$ of the observable state under evaluation increases, the action value $Q(s, \, \alpha)$ approaches the expected payoff when the rollout schemata are sampled with the probabilities equaling to their limiting frequencies of occurrence as in equation~\ref{GeiringerThmMainEq} rather rapidly unless the payoff values are truly huge in size. A pictorial diagram appears in figure\ref{IllustrateAlgorithmPic}:\footnote{Notice that we omit the initial population (or sample) $P$ when writing $Order(i \downarrow j)$ (i.e. we write $Order(i \downarrow j)$ in place of $Order(i \downarrow j, \, P)$) since the population $P$ plays no explicit role in the dynamic parallel algorithm described above. Moreover, as the new rollouts are simulated so that the weights along the digraph are incremented accordingly and potential new nodes are added, the initial population of simulated rollouts changes accordingly and yet, this does has no relevance to the algorithmic implementation.}
let's say a ``bug" has started its trip at the action $\alpha$ under evaluation. Then the bug travels to the similarity class $i_2$ with probability $\frac{Order(\alpha \downarrow i_2)}{Order(\alpha \downarrow i_2) + Order(\alpha \downarrow i_4)}$ while it travels to the node $i_4$ with probability $\frac{Order(\alpha \downarrow i_4)}{Order(\alpha \downarrow i_2) + Order(\alpha \downarrow i_4)}$. Likewise, if the bug is at the similarity class $i_5$ then it remains at $i_5$ with probability $\frac{Order(i_5 \downarrow i_5)}{Order(i_5 \downarrow i_5) + Order(i_5 \downarrow i_3)+Order(i_5 \downarrow i_8)}$, travels to $i_3$ with probability $\frac{Order(i_5 \downarrow i_3)}{Order(i_5 \downarrow i_5) + Order(i_5 \downarrow i_3)+Order(i_5 \downarrow i_8)}$ and to $i_8$ with probability $\frac{Order(i_5 \downarrow i_8)}{Order(i_5 \downarrow i_5) + Order(i_5 \downarrow i_3)+Order(i_5 \downarrow i_8)}$. Notice that if the bug ended up at the state $i_3$, it travels to the terminal label $f_1$ with probability $1$. 
\section{A Very Brief Description of \emph{Memory Evolutive Systems}: a Model of Cognition in Biological Neural Networks and Connections to the Novel Algorithms Presented in the Current Article}\label{MemoryEvolvSect}
A ``memory evolutive system" (see \cite{EhresmannVanbremeersch2007Book} for a detailed exposition by the inventors of the model) is a dynamical system of evolving multilayered digraphs with extra structure of associative composition of links (such mathematical structures are known as categories: see \cite{McLane}, or, for a gentler introduction that's more suitable to computer scientists, \cite{BarrWells}) that are meant to model evolving cognitive processes in biological neural networks. One can think of a biological neural network as a directed graph where neurons are the nodes (called objects in a category-theoretic language) and synapses from neuron $i$ to neuron $j$ are links (called morphisms or arrows, in the language of category theory). Furthermore, this category has a further extra structure: it is a weighted category in the sense that the arrows (the synapsis) have varying strengths and the strengths vary with time so that one is looking at an entire collection of weighted categories that evolves over time: $\{\mathcal{C}_t\}_{t=1}^{\infty}$. Another crucial component of the model is the hierarchy: the idea is that the concept formation is a multistage process, that takes place according to a few common computational mechanisms based on Hebbian learning i.e. the process of increasing or decreasing the strength between the synapses based on the presence or absence of signals from one neighboring neuron to another\cite{Hebb49}. Concepts form gradually as clusters of interconnected neurons that collectively influence another neuron or another cluster of interconnected neurons. Such clusters initially form \emph{cocone} diagrams and gradually ``converge" towards \emph{colimit} diagrams (see \cite{McLane}, \cite{BarrWells} or any other textbook on category theory for rigorous definitions ranging in a variety of levels of comprehension). An important reason why the notion of a colimit diagram is selected to model a conceptual cluster of neurons (or a cluster of conceptual clusters of neurons in a higher level) is that an entire collection of various arrows (in case of simple neurons synapses) corresponds uniquely to what's called a ``collective link" or collective arrow from one cluster of neurons to the other. In this manner we can think of sophisticated pre-learned concepts as objects on a higher level of development while the arrows from one such object to another are collective links representing an entire family of lower-level collective links (a collection of synapses from one neuron in the cluster to another one in the cluster when a previous level of development is the lowest one).

Returning back to the parallel algorithms described in the previous section, imagine that the conceptual clusters of neurons that are modeled as colimit diagrams represent states while collective links from these states towards other, motor-related conceptual clusters of neurons, represent actions at these states much like on the diagram in figure~\ref{ConseptDiagrP} below.  
\begin{figure}[htp]
\centerline{\includegraphics[height=10cm]{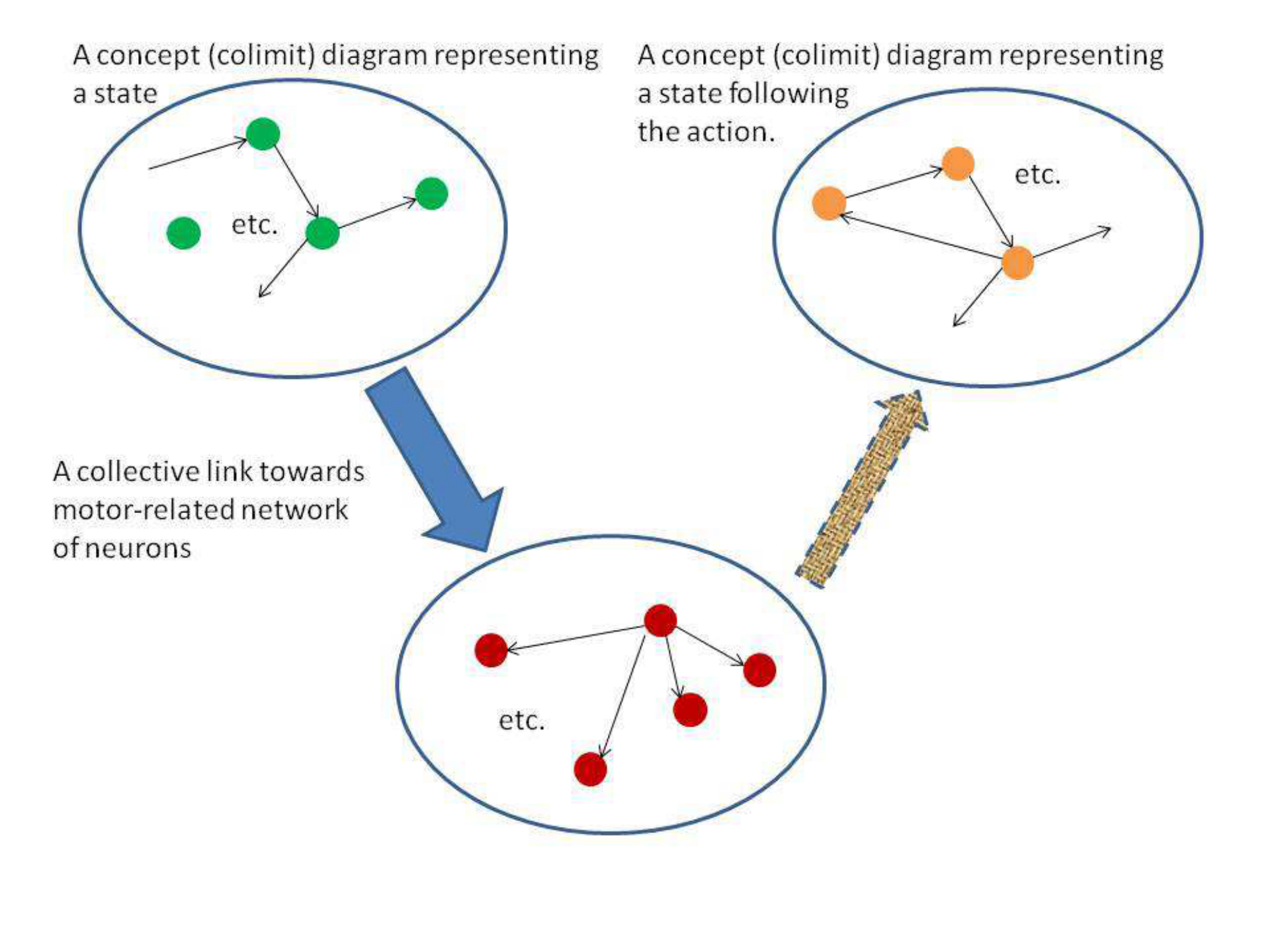}}
\caption{A conceptual perception of the new state represented by a cluster of neurons on the right followed after an execution of an action available at the state represented by a cluster of neurons on the left.}
\label{ConseptDiagrP}
\end{figure}
Eventually, a state-action sequence leads to a state that has a numerical payoff measured by the intensity of a feedback signal thereby resembling the notion of a rollout in Monte-Carlo 
tree search and POMDPs. It is conceivable then that the action evaluation mechanisms for the collective links towards the motor-related clusters of neurons are implemented in a similar manner as described at the end of the previous section. Moreover, the newest generalized Geiringer-like theorem in \cite{MitavJunGeirFoga2013} that allows recombination over arbitrary set cover, leads to new fascinating insights into payoff-based clustering algorithms that resemble self-organizing maps and may also play a significant role in the concept-formation. These algorithms are the subject of sequel papers.
\section{Conclusions and Future Research Directions}\label{conclusionsSect}
In the most general setting, Geiringer-like theorems address the limiting frequency of occurrence of various generalized ``genetic material" that ranges from alleles in genomic expressions to functions in genetic programming or, even more interestingly, observable states in computational intelligence in the long run after repeated application of recombination (or crossover) transformations. In the current paper we describe a rather simple dynamic algorithm motivated by one of the latest theorems in \cite{MitavRowGeirNewMain} to take advantage of randomness and incomplete information by predicting the outcome of the limiting frequency of occurrence of various states after an ``infinitely long" recombination process as described in section~\ref{FrameworkAndAlgorithmSect}. Not only such algorithms are anticipated to have a tremendous impact when coping with POMDPs (partially observable Markov decision processes), a link has been exhibited between such algorithms for decision making in the environments with randomness and incomplete information and the corresponding algorithms in biological neural networks through an elegant and well-developed category-theoretic model of cognition in \cite{EhresmannVanbremeersch2007Book}. The most recent version of the theorem in \cite{MitavJunGeirFoga2013} motivates further extensions of the dynamic algorithms in the current paper for payoff-based clustering and will be investigated in the sequel papers.


\section*{Acknowledgements}
This work has been supported by the EPSRC EP/I009809/1 ``Evolutionary Approximation Algorithms for Optimization: Algorithm Design and Complexity Analysis" Grant.

\end{document}